\documentclass[runningheads]{llncs}
\usepackage{graphicx}

\usepackage{tikz}
\usepackage{comment}
\usepackage{amsmath,amssymb} %
\usepackage{color}
\usepackage{graphicx}
\usepackage{amsmath}
\usepackage{amssymb}
\usepackage{booktabs}
\usepackage{algorithm}
\usepackage{algpseudocode}
\usepackage{multirow}
\usepackage{subfig}
\usepackage{hyperref}

\newcommand{\etal}{\textit{et al.~}}

\begin{document}
\pagestyle{headings}
\mainmatter
\def\ECCVSubNumber{2311}  %

\title{SESS: Saliency Enhancing with Scaling and Sliding} %

\titlerunning{Abbreviated paper title}
\author{Osman Tursun  \and
Simon Denman \and Sridha Sridharan \and
Clinton Fookes}
\authorrunning{O. Tursun et al.}
\institute{SAIVT Lab, Queensland University of Technology, Australia
\\
\email{\{osman.tursun,s.denman,s.sridharan,c.fookes\}@qut.edu.au}
}
\maketitle

\begin{abstract}
High-quality saliency maps are essential in several machine learning application areas including explainable AI and weakly supervised object detection and segmentation. Many techniques have been developed to generate better saliency using neural networks. However, they are often limited to specific saliency visualisation methods or saliency issues. We propose a novel saliency enhancing approach called \textbf{SESS} (\textbf{S}aliency \textbf{E}nhancing with \textbf{S}caling and \textbf{S}liding). It is a method and model agnostic extension to existing saliency map generation methods. With SESS, existing saliency approaches become robust to scale variance, multiple occurrences of target objects, presence of distractors and generate less noisy and more discriminative saliency maps. SESS improves saliency by fusing saliency maps extracted from multiple patches at different scales from different areas, and combines these individual maps using a novel fusion scheme that incorporates channel-wise weights and spatial weighted average. To improve efficiency, we introduce a pre-filtering step that can exclude uninformative saliency maps to improve efficiency while still enhancing overall results. We evaluate SESS on object recognition and detection benchmarks where it achieves significant improvement. The code is released publicly to enable researchers to verify performance and further development. Code is available at:\url{https://github.com/neouyghur/SESS}

\end{abstract}

\section{Introduction}
\label{sec:intro}

Approaches that generate saliency or importance maps based on the decision of deep neural networks (DNNs) are critical in several machine learning application areas including  explainable AI and weakly supervised object detection and semantic segmentation. High-quality saliency maps increase the understanding and interpretability of a DNN's decision-making process, and can increase the accuracy of segmentation and detection results.

Since the development of DNNs, numerous approaches have been proposed to efficiently produce high-quality saliency maps. However, most methods have limited transferability and versatility. Existing methods are designed for DNN models with specific structures (i.e. a global average pooling layer), for certain types of visualisation (for details refer to Sec. \ref{sec:lit}), or to address a specific limitation. For instance, CAM \cite{zhou2016learning} requires a network with global average pooling. Guided backpropagation (Guided-BP) \cite{springenberg2014striving} is restricted to gradient-based approaches. Score-CAM \cite{wang2020score} seeks to reduce the method's running-time, while SmoothGrad \cite{smilkov2017smoothgrad} aims to generate saliency maps with lower noise.

\begin{figure*}[!th]
	\centering
	\includegraphics[width=\linewidth]{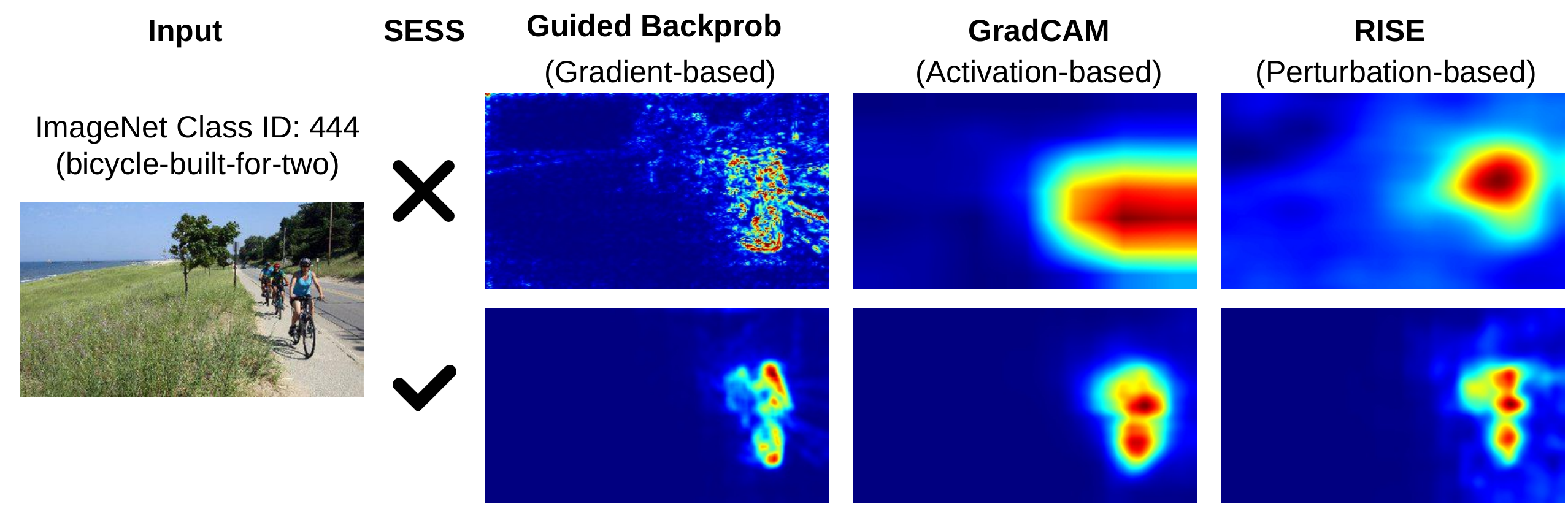}
	\caption{Example results of three well-known deep neural network visualisation methods with and without SESS. Each of these methods represents one type of saliency map extraction technique. With SESS, all methods generate less noisy and more discriminative saliency maps. The results are extracted with ResNet50, and layer4 is used for Grad-CAM. Target ImageNet class ID is 444 (bicycle-built-for-two).}
	\label{fig:mov}
\end{figure*}

In this work, we propose Saliency Enhancing with Scaling and Sliding (SESS), a model and method agnostic black-box extension to existing saliency visualisation approaches. SESS is only applied to the input and output spaces, and thus does not need to access the internal structure and features of DNNs, and is not sensitive to the design of the base saliency method. It also addresses multiple limitations that plague existing saliency methods. For example, in Fig. \ref{fig:mov}, SESS shows improvements when applied to three different saliency methods. The saliency map extracted with the gradient-based approach (Guided-BP) is discriminative but noisy. Saliency maps generated by the activation-based method Grad-CAM \cite{selvaraju2017grad} and perturbation-based method RISE \cite{petsiuk2018rise} generate smooth saliency maps, but lack detail around the target object and fail to precisely separate the target object from the scene. With SESS, the results of all three methods become less noisy and a more discriminative boundary around the target is obtained.

SESS addresses the following limitations of existing approaches:
\begin{itemize}
	\item {\bf{Weak scale invariance:}} Several studies claim that generated saliency maps are inconsistent when there are scale differences \cite{jo2021puzzle,wang2020self}, and we also observe that generated saliency maps are less discriminative when the target objects are comparatively small (See Fig. \ref{fig:mov} and Fig. \ref{fig:qual}).
	\item {\bf{Inability to detect multiple occurrences:}} Some deep visualisation methods (i.e., Grad-CAM) fail to capture multiple occurrences of the same object in a scene \cite{chattopadhay2018grad,omeiza2019smooth}. (See Fig. \ref{fig:qual}).
	\item {\bf{Impacted by distractors:}} Extracted saliency maps frequently incorrectly highlight regions when distractors exist. This is especially true when the class of the distractor returns a high confidence score, or is correlated with the target class.
	\item {\bf{Noisy results:}} Saliency maps extracted with gradient based visualisation approaches  \cite{simonyan2013deep,sundararajan2017axiomatic} appear visually noisy as shown in Fig \ref{fig:mov}. 
	\item {\bf{Less discriminative results:}} Activation based approaches (e.g., Grad-CAM) tend to be less discriminative, often highlighting large regions around the target such that background regions are often incorrectly captured as being salient.
	\item {\bf{Fixed input size requirements:}} Neural networks with fully-connected layers like VGG-16 \cite{simonyan2014very} require a fixed input size. Moreover, models perform better when the input size at inference is the same as the input size during training. As such, most visualisation methods resize the input to a fixed size. This impacts the resolution and aspect ratio, and may cause poor visualisation results \cite{wang2020self}.
\end{itemize}

SESS is a remedy for all of the limitations mentioned above. SESS extracts multiple equally sized (i.e., $224 \times 224$) patches from different regions of multiple scaled versions of an input image through resizing and sliding window operations. This step ensures that it is robust to scale variance and multiple occurrences.
Moreover, since each extracted patch is equal in size to the default input size of the model, SESS takes advantage of high-resolution inputs and respects the aspect ratio of the input image. Each extracted patch will contribute to the final saliency map, and the final saliency map is the fusion of the saliency maps extracted from patches. In the fusion step, SESS considers the confidence score of each patch, which serves to reduce noise and the impact of distractors while increasing SESS's discriminative power. 

The increased performance of SESS is achieved countered a reduction in efficiency due to the use of multiple patches. Quantitative ablation studies show using more scales and denser sliding windows are beneficial, but increase computational costs. To reduce this cost, SESS uses a pre-filtering step that filters out background regions with low target class activation scores. Compared to saliency extraction, the inference step is efficient as it only requires a single forward pass and can exploit parallel computation and batch processing. As such, SESS obtains improved saliency masks with a small increase in run-time requirements. Ablation studies show that the proposed method outperforms its base saliency methods when using pre-filtering with a high pre-filter ratio. In a Pointing Game experiment \cite{zhang2018top} all methods with SESS achieved significant improvements, despite of a pre-filter ratio of $99\%$ that excludes the majority of extracted patches from saliency generation.

We quantitatively and qualitatively evaluate SESS and conduct ablation studies regarding multiple scales, pre-filtering and fusion. All experimental results show that SESS is a useful and versatile extension to existing saliency methods.

To summarize, the main contributions of this work are as follows:

\begin{itemize}
	\item We propose, SESS, a model and method agnostic black-box extension to existing saliency methods which is simple and efficient.
	\item  We demonstrate that SESS increases the visual quality of saliency maps, and improves their performance on object recognition and localisation tasks.
\end{itemize}
\section{Related Work}
\label{sec:lit}
{\noindent \bfseries{Deep Saliency Methods:}} Numerous deep neural network-based visualisation methods have been developed in recent years. Based on how the saliency map is extracted, they can be broadly categorised into three groups: gradient-based \cite{simonyan2013deep,smilkov2017smoothgrad,sundararajan2017axiomatic}, class activation-based\cite{zhou2016learning,selvaraju2017grad,zhang2021group,wang2020score}, and perturbation-based \cite{fong2017interpretable,petsiuk2018rise,dabkowski2017real} methods.

Gradient-based methods interpret the gradient with respect to the input image as a saliency map. They are efficient as they only require a single forward and backward propagation operation. However, saliency maps generated from raw gradients are visually noisy. Activation-based methods aggregate target class activations of a selected network layer to generate saliency maps. Compared with gradient-based methods, activation-based methods are less noisy, but are also less discriminative and will often incorrectly show strong activations in nearby background regions. Perturbation-based methods generate saliency maps by measuring the changes in the output when the input is perturbed. Perturbation-based methods are slow when compared to most gradient- and activation-based approaches, as they require multiple queries.

Methods can also be split into black-box and white-box according to whether they access the model architecture and parameters. Except for some perturbation-based methods \cite{petsiuk2018rise,fong2017interpretable}, saliency methods are all white-box in nature \cite{simonyan2013deep,smilkov2017smoothgrad,sundararajan2017axiomatic,zhou2016learning,selvaraju2017grad}. White-box methods are usually more computationally efficient than black-box methods, and require a single forward and backward pass through the network. However, black-box methods are model agnostic, while white-box methods may only work on models with specific architectural features.

Approaches can also be one-shot or multi-shot in nature. One-shot approaches require a single forward and backward pass. Most gradient- and activation-based methods are single-shot. However, multi-shot variants are developed to obtain further improvements. For example, SmoothGrad \cite{smilkov2017smoothgrad} generates a sharper visualisations through multiple passes across noisy samples of the input image. Integrated Gradients (IG) \cite{sundararajan2017axiomatic} addresses the ``gradient saturation" problem by taking the average of the gradients across multiple interpolated images. Augmented Grad-CAM \cite{morbidelli2020augmented} generates high-resolution saliency maps through multiple low-resolution saliency maps extracted from augmented variants of the input. Smooth Grad-CAM++ \cite{omeiza2019smooth} utilises the same idea proposed in SmoothGrad to generate sharper saliency maps. To the best of our knowledge, all perturbation methods are multi-shot in nature, as they require multiple queries to the model, each of which has a different perturbation. 

Attempts have been made to make multi-shot approaches more efficient. Most such approaches seek to create perturbation masks in an efficient way. Dabkowski \etal \cite{dabkowski2017real} generates a perturbation mask with a second neural network. Score-CAM \cite{wang2020score} uses class activation maps (CAM) as masks; and Group-CAM \cite{zhang2021group} follows a similar idea, but further reduces the number of masks through the merging of adjacent maps.

The proposed SESS is a method and model agnostic saliency extension. It can be a ``plug-and-play" extension for any saliency methods. However, like perturbation methods, it requires multiple queries. As such, for the sake of efficiency, single-shot and efficient multi-shot approaches are most appropriate for use with SESS.

{\noindent \bfseries{Enhancing Deep Saliency Visualisations:}} Many attempts have been made to generate discriminative and low-noise saliency maps. Early Gradient-based methods are visually noisy, and several methods have been proposed to address this. Guided-BP \cite{springenberg2014striving} ignores zero gradients during backpropagation by using a RELU as the activation unit. SmoothGrad \cite{springenberg2014striving} takes the average gradient of noisy samples \cite{smilkov2017smoothgrad} to generate cleaner results.

The first of the  activation-based methods, CAM, is model sensitive. It requires the model apply a global average pooling over convolutional feature map channels immediately prior to the classification layer \cite{petsiuk2018rise}. Later variants such as Grad-CAM relax this restriction by using average channel gradients as weights. However, Grad-CAM \cite{selvaraju2017grad} is also less discriminative, and is unable to locate multiple occurrences of target objects. Grad-CAM++ \cite{chattopadhay2018grad} uses positive partial derivatives of features maps as weights. Smooth Grad-CAM++ \cite{omeiza2019smooth} combines techniques from both Grad-CAM++ and SmoothGrad to generate sharper visualisations.

Perturbation methods are inefficient, as they send multiple queries to the model. For example, RISE \cite{petsiuk2018rise} sends 8000 queries to the model to evaluate the importance of regions covered by 8000 randomly selected masks. Recent works reduce the number of masks by using channels in CAMs as masks. For instance, Score-CAM uses all channels in CAMs, while Group-CAM further minimises the number of masks by grouping the channels of CAMs.

All the aforementioned methods have successfully improved certain issues relating to saliency methods, but have limited transferability and versatility. In comparison, SESS is a model and method agnostic extension, which can be applied to any existing saliency approach (though we note that single-pass or efficient multi-pass methods are most suitable). Moreover, SESS is robust to scale-variance, noise, multiple occurrences and distractors. SESS can generate clean and focused saliency maps, and significantly improves the performance of saliency methods for image recognition and detection tasks.

\section{Saliency Enhancing with Scaling and Sliding (SESS)}
\label{sec:method}
\begin{figure*}[t]
	\centering
	\includegraphics[width=\linewidth]{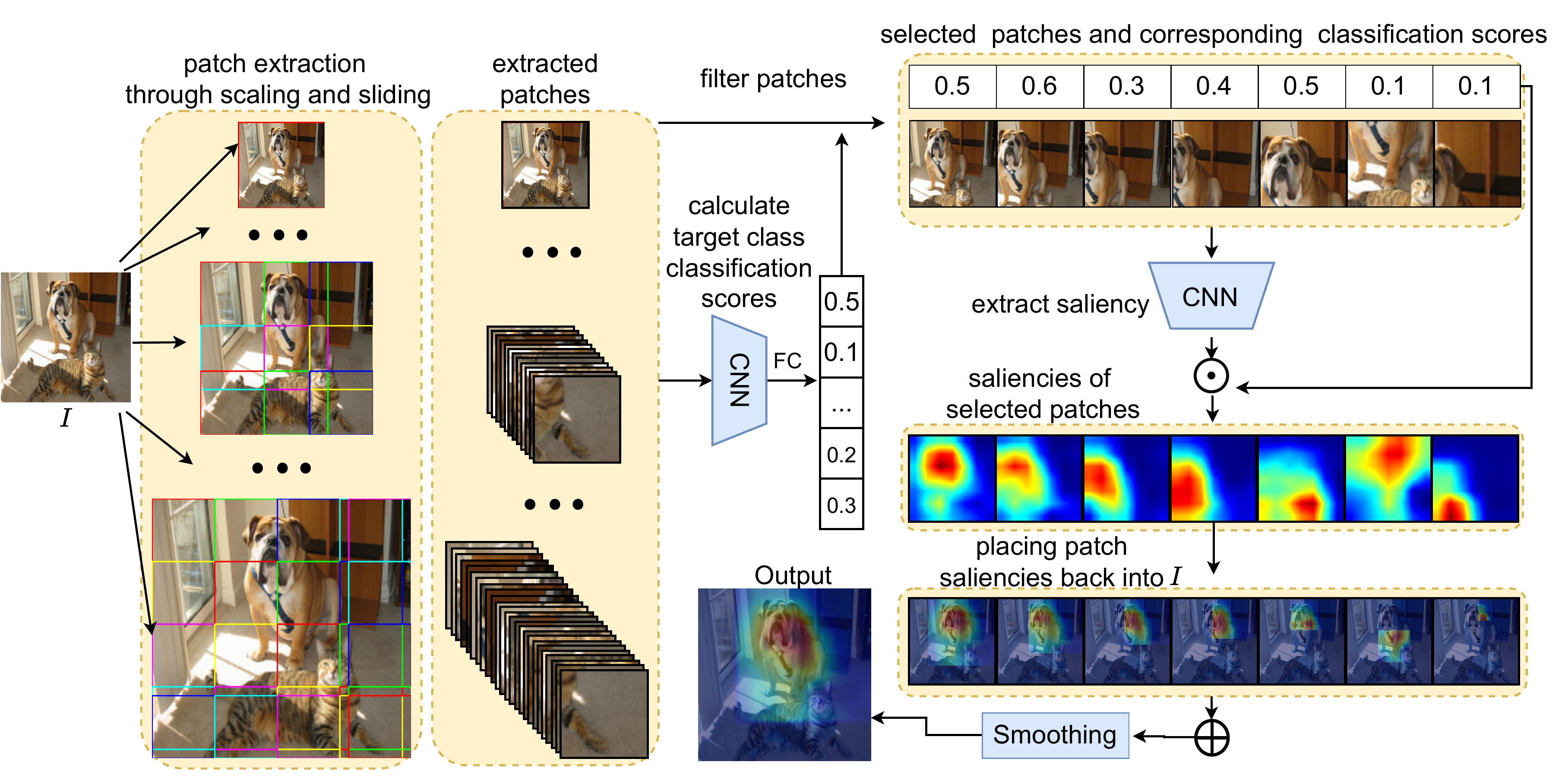}
	\caption{The SESS Process: SESS includes six major steps: multi-scaling, sliding window, pre-filtering, saliency extraction, saliency fusion and smoothing.}
	\label{fig:sess}
\end{figure*}

In this section, we introduce SESS. A system diagram is shown in Figure \ref{fig:sess}, and the main steps are described in Algorithm \ref{algo:sess}. The implementation of SESS is simple and includes six steps: \textit{multi-scaling, sliding window, pre-filtering, saliency extraction, saliency fusion, and smoothing}. The first four steps are applied to the input space, and the last two steps are applied at the output space. SESS is therefore a black-box extension and a model and method agnostic approach. Each of these steps will be discussed in detail in this section.
\renewcommand{\algorithmicrequire}{\textbf{Input:}}
\renewcommand{\algorithmicensure}{\textbf{Output:}}
\begin{algorithm}[!ht]
	\caption{SESS}
	\label{algo:sess}
	\begin{algorithmic}[1]
		\Require Image $I$, Model $f$, Target class $c$, Scale n, Window size $(w, h)$, Pre-filtering ratio $r$
		\Ensure Saliency map $L^c_{sess}$
		\State $M, P \gets []$
		\For{$i \in [1, \dots, n]$}
		\Comment{Scaling}
		\State $M$.append(resize(I, 224 + 64 $\times$ $(i - 1)$))
		\EndFor
		
		\For{$m \in M$}
		\Comment{Extracting patches}
		\State $P$.append(sliding-window(m, w, h))
		\EndFor
		
		\State $B$ $\gets$ batchify($P$)
		\State $S^c$ $\gets$ $f(B, c)$
		\Comment{$S^c$ as activation scores of class $c$ }
		\State $S_{fil.}^c, P_{fil.} \gets$ pre-filtering($S^c$, $P$, $r$)
		\Comment{filter out patches whose class $c$ activation score is lower than top $(100 - r)\%$}
		\State $A$ $\gets$ saliency\_extraction($P_{fil.}$, $f$, $c$) 
		\Comment{get saliency maps of patches after pre-filtering}
		\State $L$ $\gets$ calibration($P_{fil.}$, $A$) \Comment{$L$ is a tensor with shape $n \times w \times h$ }
		\State $L' \gets L \otimes S_{fil.}^c$ \Comment{Apply channel-wise weight}
		\State $L^c_{sess} =$ weighted\_average($L'$) \Comment{Apply binary weights to obtain the average of the non-zero values}
		\State \Return $L^c_{sess}$
	\end{algorithmic}
\end{algorithm}

\noindent{\bfseries{Multi-scaling:}} Generating multiple scaled versions of the input image $I$ is the first step of SESS. In this study, the number of scales, $n$, ranges from 1 to 12. The set of sizes of all scales is equal to $\{224 + 64*(i-1) | i \in \{1, 2, \dots,n\}\}$.  The smallest size is equal to the default size of pre-trained models, and the largest size is approximately four times the smallest size. The smaller side of $I$ is resized to the given scale, while respecting the original aspect ratio. $M$ represents the set of all $I$s at different scales.
Benefits of multi-scaling include:
\begin{itemize}
	\item Most saliency extraction methods are scale-variant. Thus saliency maps generated at  different scales are inconsistent. By using multiple scales and combining the saliency results from these, scale-invariance is achievable.
	\item Small objects will be distinct and visible in salience maps after scaling. 
\end{itemize}
\newcommand{\ceil}[1]{\lceil {#1} \rceil}
\noindent{\bfseries{Sliding window:}} For efficiency the sliding window step occurs after multi-scaling, which calls $n$ resizing operations. A sliding window is applied to each image in $M$ to extract patches. The width $w$ and height $h$ of the sliding window is set to $224$. Thus patch sizes are equal to the default input size of pre-trained models in PyTorch\footnote{https://pytorch.org}. The sliding operation starts from the top-left corner of the given image, and slides through from top to bottom and left to right. By default, for efficiency, the step-size of the sliding window is set to 224, in other words there is no overlap between neighbouring windows. However, patches at image boundaries are allowed to overlap with their neighbours to ensure that the entire image is sampled. The minimum number of generated patches is $\sum_{i=1}^{n}\ceil{0.25i +0.75}^2$. When $I$ has equal width and height and $n=1$, only one patch of size $224 \times 224$ will be extracted, and SESS will return the same results as it's base saliency visualisation method. Thus, SESS can be viewed as a generalisation of existing saliency extraction methods.

\begin{figure*}[!th]
	\centering
	\includegraphics[width=\linewidth]{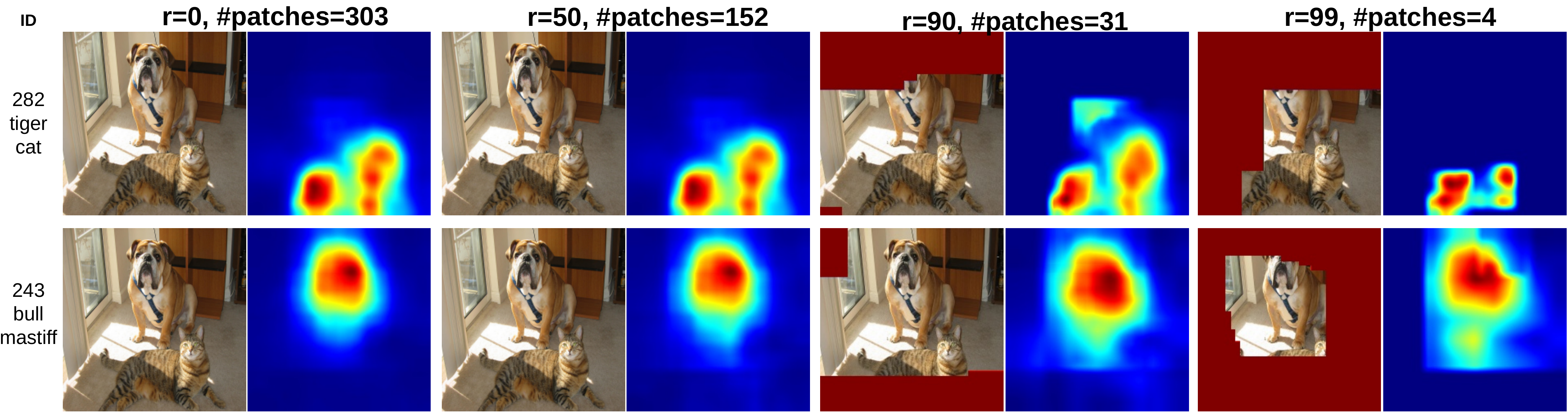}
	\caption{Visualisation of regions and saliency maps after pre-filtering, when computing saliency maps for the target classes ``tiger cat" (top row) and ``bull mastiff" (bottom row). All patches that overlap with the red region are removed after pre-filtering.}
	\label{fig:filter}
\end{figure*}

\noindent{\bfseries{Pre-filtering:}} To increase the efficiency of SESS, a pre-filtering step is introduced. Generating saliency maps for each extracted patch is computationally expensive. Generally, only a few patches are extracted that contain objects which belong to the target class, and they have comparatively large target class activation scores. Calculating target class activation scores requires only a forward pass, and can be sped-up by exploiting batch operations. After sorting the patches based on activation scores, only patches that have a score in the top $(100 - r)\%$ of patches are selected to generate saliency maps. Here, we denote $r$ the pre-filter ratio. When $r=0$, no pre-filter is applied. As shown in Fig. \ref{fig:filter}, when $r$ increases only the region which covers the target object remains, and the number of patches is greatly reduced. For instance, only four patches from an initial set of 303 patches are retained after applying a pre-filter with $r=99$, and these patches are exclusively focussed on the target object. Of course, a large pre-filter ratio i.e., $r > 50$ will decrease the quality of the generated saliency maps as shown in Fig. 
\ref{fig:filter}. Note we use notation $S^c_{fil}$ to represent the class ``c" activation scores of the remaining patches after filtering.

\noindent{\bfseries{Saliency extraction:}} The saliency maps for the patches retained after pre-filtering are extracted with a base saliency extraction method. Any saliency extraction method is suitable; however real-time saliency extraction methods including Grad-CAM, Guided-BP and Group-CAM are recommended for efficiency. Each extracted saliency map is normalised with Min-Max normalisation.%

\noindent{\bfseries{Saliency fusion:}} Since each patch is extracted from a different position or a scaled version of $I$, a calibration step is applied before fusion. Each saliency map is overlayed on a zero mask image which has the same size as the scaled $I$ from which it was extracted. Then all masks are resized to the same size as $I$. Here, notation $L$ represents the channel-wise concatenation of all masks. $L$ has $n$ channels of size $w \times h$. Before fusion, a channel-wise weight is applied. $S^c_{fil}$, the activation scores of patches after filtering, is used as the weight. The weighted $L'$ is then obtained using,

\begin{equation}¬
	L' = L \otimes S^c_{fil}. 
\end{equation}
 
Finally, a weighted average that excludes non-zero values is applied at each spatial position for fusion. The modified weighted average is used over uniform average to ignore the zero saliency values introduced during the calibration step. Thus, the saliency value at $(i, j)$ of the final saliency map becomes, 

\begin{equation}
	L_{sess}(i,j) = \frac{\sum_{i=1}^{n}L'(n,i, j)*\sigma(L'(n,i, j))}{\sum_{i=1}^{n}\sigma(L'(n,i, j))},
\end{equation}

where $\sigma(x)=1$ if $x>\theta$, else $\sigma(x)=0$, and $\theta = 0$. A Min-Max normalisation is applied after fusion.

\begin{figure}[!htb]
	\centering
	\subfloat[input]{\includegraphics[width=0.25\linewidth]{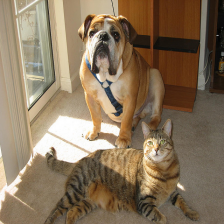}}\quad
	\subfloat[before smoothing]{\includegraphics[width=0.25\linewidth]{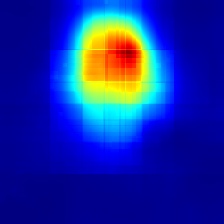}}\quad
	\subfloat[after smoothing]{\includegraphics[width=0.25\linewidth]{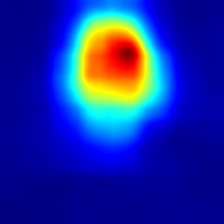}}
	\caption{An example of the effect of the smoothing step. After the smoothing step, edge artefacts are removed and the generated saliency is more visually pleasing.}
	\label{fig:smooth}
\end{figure}

\noindent{\bfseries{Smoothing:}} Visual artefacts typically remain between patches after fusion, as shown in Fig. \ref{fig:smooth}. Gaussian filtering is applied to eliminate these artefacts. This paper sets the kernel size to 11 and $\sigma$ = 5.

\section{Experiments}
\label{sec:exp}
In this section, we first conduct a series of ablation studies to find the optimal hyper-parameters and show the significance of the steps in our approach. Then, we qualitatively and quantitatively evaluate the efficiency and effectiveness of SESS compared to other widely used saliency methods.

\subsection{Experimental Setup}
All experiments are conducted on the validation split of the three publicly available datasets: ImageNet-1k \cite{russakovsky2015imagenet}, PASCAL VOC07 \cite{everingham2010pascal} and MSCOCO2014 \cite{lin2014microsoft}. Pre-trained VGG-16 (layer: Feature.29) \cite{simonyan2014very} and ResNet-50 (layer: layer4)
\cite{he2016deep} networks are used as backbones in our experiments. We used Grad-CAM \cite{selvaraju2017grad}, Guided-BP \cite{springenberg2014striving} and Group-CAM \cite{zhang2021group} as base saliency extraction methods. Grad-CAM and Guided-BP are selected as widely used representations of activation-based and gradient-based approaches. We selected Group-CAM as a representative perturbation-based method given it's efficiency.

In qualitative experiments, the number of scales and the pre-filter ratio are set to 12 and 0, and smoothing is applied. In quantitative experiments, we employ fewer scales and higher pre-filtering ratios, and omit the smoothing step.

\subsection{Ablation Studies}
We conduct ablation studies on 2000 random images selected from the validation split of ImageNet-1k \cite{russakovsky2015imagenet}. %
ImageNet pre-trained VGG-16 and ResNet-50 
networks are used during the ablation study, and Grad-CAM is used as the base saliency method. Insertion and deletion scores \cite{petsiuk2018rise} are used as evaluation metrics. The intuition behind this metric is that deletion/insertion of pixels with high saliency will cause a sharp drop/increase in the classification score of the target class. The area under the classification score curve (AUC) is used as the quantitative indicator of the insertion/deletion score. A lower deletion score and a higher insertion score indicates a high-quality saliency map. We also reported the overall score as in \cite{zhang2021group}, where the overall score is defined as $AUC(insertion) - AUC(deletion)$. The implementation is the same as \cite{zhang2021group}. $3.6\%$ of pixels are gradually deleted from the original image in the deletion test, while $3.6\%$ of pixels are recovered from a highly blurred version of the original image in the insertion test.    

\begin{figure*}[!ht]
	\centering
	\subfloat[][Insertion curve]{\includegraphics[width=0.3\linewidth]{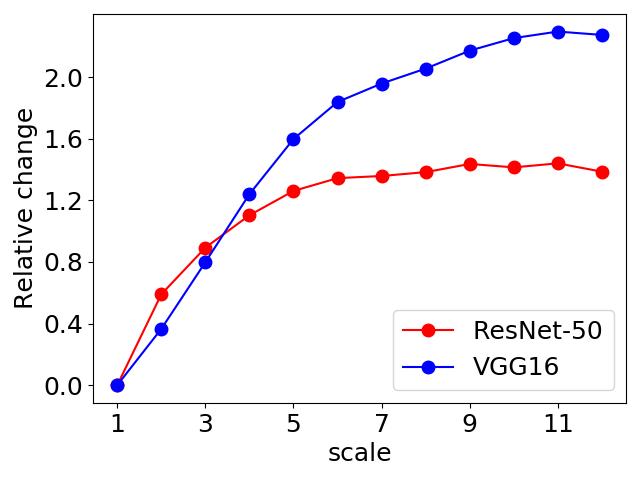}\label{fig:sc_ins}}
	\subfloat[][Deletion curve]{\includegraphics[width=0.3\linewidth]{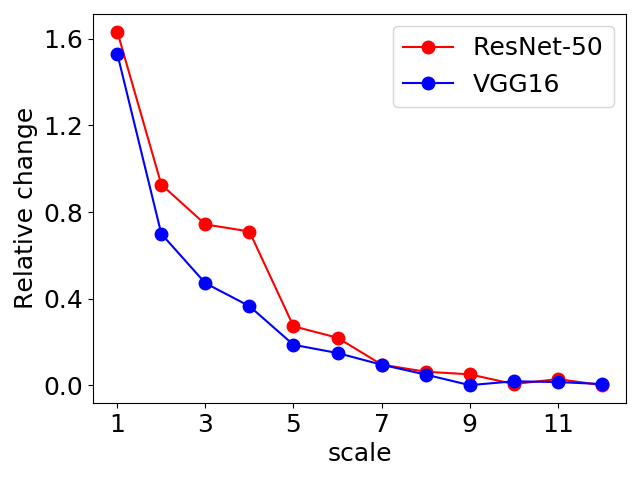}\label{fig:sc_del}}
	\subfloat[][Overall curve]{\includegraphics[width=0.3\linewidth]{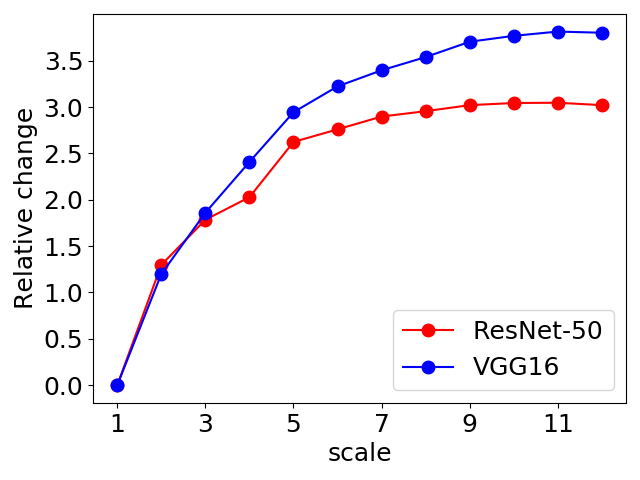}\label{fig:sc_overall}}
	\caption{Ablation study considering scale factor in terms of deletion (lower AUC is better), insertion (higher AUC is better), and overall score (higher AUC is better) on the ImageNet-1k validation split (on a randomly selected set of 2k images).}
	\label{fig:scale}
\end{figure*}

\noindent{\bfseries{Scale:}} To study the role of multi-scale inputs, we tested different numbers of scales with insertion and deletion tests. As Figure \ref{fig:scale} shows, for both VGG-16 and ResNet-50, when the number of scales increases, the insertion scores increase and the deletion scores decrease. Improvements begin to plateau once five scales are used, and converge once ten scales are used. Overall, the improvement is clear even when using images from the ImageNet dataset, where the main object typically covers the majority of the image, and the role of scaling is less apparent.

\noindent{\bfseries{Pre-filtering ratio:}} To find a high pre-filtering ratio which increases efficiency whilst retaining high performance, we test 10 different global filters from 0 to 0.9. The insertion score decreases as the pre-filter ratio increases, while the deletion scores fluctuate just slightly until the pre-filter ratio reaches 0.6, after which they increase sharply. This shows pre-filter ratio can be set to 0.5 for both high quality and efficiency. However, we used a pre-filter ratio larger than 0.9 in quantitative experiments.

\noindent{\bfseries{Channel-wise weights:}} In the fusion step of SESS, channel-wise weights are applied. Figure \ref{fig:wa} qualitatively shows the role of the channel-wise weights. With the channel-wise weights, the extracted saliency maps are more discriminative, better highlighting relevant image regions. Without the channel-wise weights, background regions are more likely to be detected as salient.

\begin{figure*}[!ht]
	\centering
	\subfloat[][Insertion curve]{\includegraphics[width=0.3\linewidth]{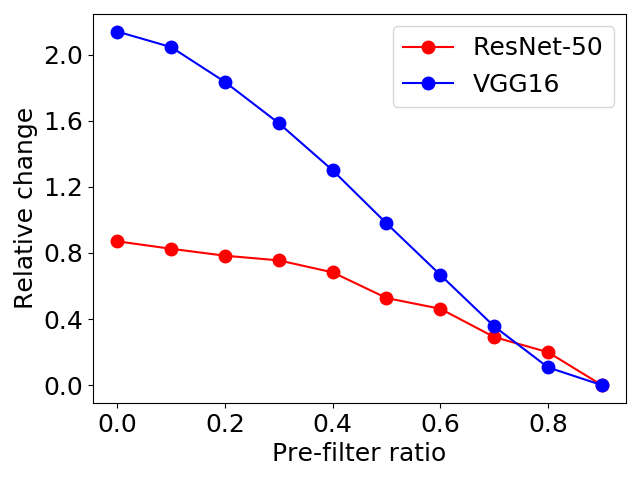}\label{fig:gt_ins}}
	\subfloat[][Deletion curve]{\includegraphics[width=0.3\linewidth]{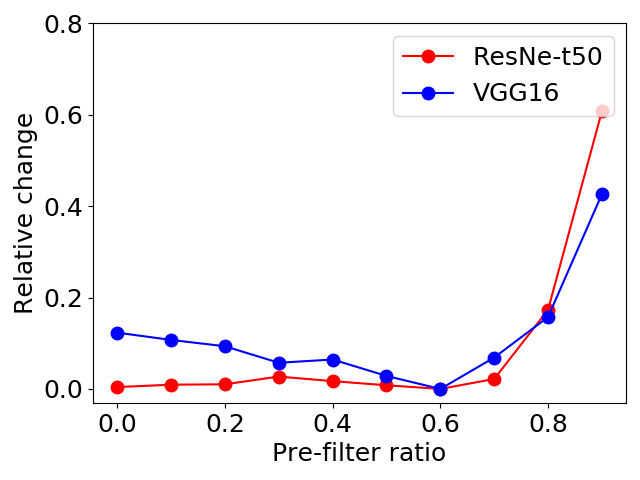}\label{fig:gt_del}}
	\subfloat[][Overall curve]{\includegraphics[width=0.3\linewidth]{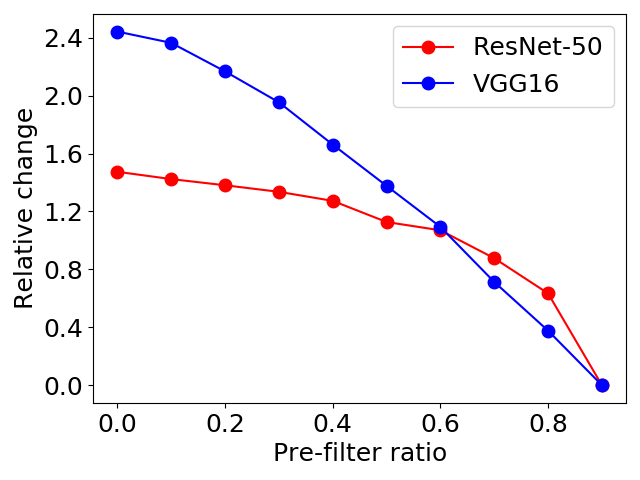}\label{fig:gt_overall}}
	\caption{Ablation study considering the pre-filtering operation in terms of deletion (lower is better), insertion (higher is better), and overall (higher is better) scores on the ImageNet-1k validation split (on a randomly selected set of 2k images).}
	\label{fig:gt}
\end{figure*}

\begin{figure*}[!ht]
	\centering
	\includegraphics[width=0.95\textwidth]{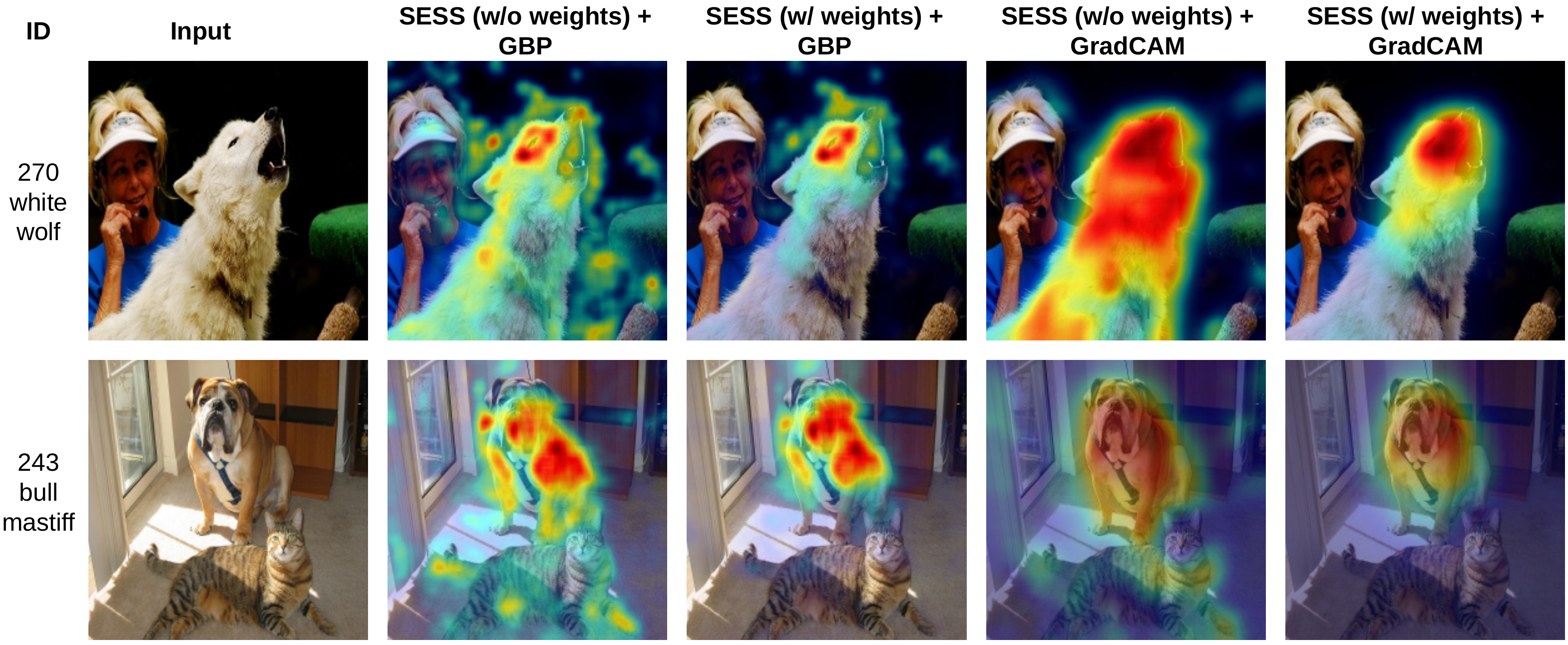}
	\caption{Impact of the channel-wise weights: The use of the channel-wise weights suppresses activations in background regions, and results in a more focussed saliency map.}
	\label{fig:wa}
\end{figure*}

\begin{table}[]
	\centering
	\begin{tabular}{c|c|c|c|c|c}
		\hline
		\multicolumn{1}{c|}{\bf Method}            & \multicolumn{1}{c|}{\bf Model/layer} & \bf SESS & \bf Insertion ($\uparrow$)    & \bf Deletion ($\downarrow$)   & \bf Over-all ($\uparrow$) \\ \hline
		\multirow{4}{*}{Grad-CAM \cite{selvaraju2017grad}}  & \multirow{2}{*}{ResNet-50}  &   & 68.1 & 12.1 & 56.0 \\ \cline{3-6} 
		&                                   &  \checkmark & 68.6 & 11.3 & 57.3  \\ \cline{2-6} 
		& \multirow{2}{*}{VGG-16} &   & 60.6  & 9.1   & 51.5  \\ \cline{3-6} 
		&                                   &  \checkmark & 60.3  & 8.1   & 52.2  \\ \hline
		\multirow{4}{*}{Guided-BP \cite{springenberg2014striving}} & \multirow{2}{*}{ResNet-50} &      & 47.8  & 11.0  & 36.8   \\ \cline{3-6} 
		&                                   &  \checkmark & 53.0  & 12.0  & 41.0  \\ \cline{2-6} 
		& \multirow{2}{*}{VGG-16} &   & 38.8  & 6.8   & 32.0  \\ \cline{3-6} 
		&                                   &  \checkmark & 44.3  & 6.9   & 37.4  \\ \hline
		\multirow{4}{*}{Group-CAM \cite{zhang2021group}} & \multirow{2}{*}{ResNet-50}  &   & 68.2  & 12.1  & 56.2  \\ \cline{3-6} 
		&                                   &  \checkmark & 68.8  & 11.3  & 57.4  \\ \cline{2-6} 
		& \multirow{2}{*}{VGG-16} &   & 61.1  & 8.8   & 52.3  \\ \cline{3-6} 
		&                                   &  \checkmark & 61.1  & 8.1   & 53.1  \\ \hline
	\end{tabular}
	\caption{Comparison in terms of deletion (lower is better), insertion (higher is better), and the overall (higher is better) scores on a randomly selected set of 5000 images from the ImageNet-1k validation split.}
	\label{tab:quan1}
\end{table}

\subsection{Image Recognition Evaluation}
Here, we also use the insertion and deletion metrics to evaluate the performance of the proposed SESS. We evaluate SESS with three base saliency extraction approaches (Grad-CAM, Guided-BP and Group-CAM) and two backbones (VGG-16 and ResNet-50) on 5000 randomly selected images from the ImageNet-1k validation split. Considering the efficiency, the number of scales and pre-filter ratio are set to 10 and 0.9. With SESS, all three methods with two different backbones achieve improvements, especially Guided-BP whose overall score increases by nearly $5\%$.

\subsection{Qualitative Results}
This section qualitatively illustrates how much visual improvement SESS brings over the base visualisation approaches such as Grad-CAM \cite{selvaraju2017grad} and Guided-BP \cite{springenberg2014striving}. As a baseline, we selected five visualisation approaches: Guided-BP, SmoothGrad \cite{smilkov2017smoothgrad}, RISE \cite{petsiuk2018rise}, Score-CAM \cite{wang2020score} and Grad-CAM. ResNet-50 is selected as the backbone for all methods. We selected challenging cases for queries, including instances with multiple occurrences of the target classes, the presence of distractors, small targets, and curved shapes.

\begin{figure*}[!ht]
	\centering
	\includegraphics[width=\textwidth]{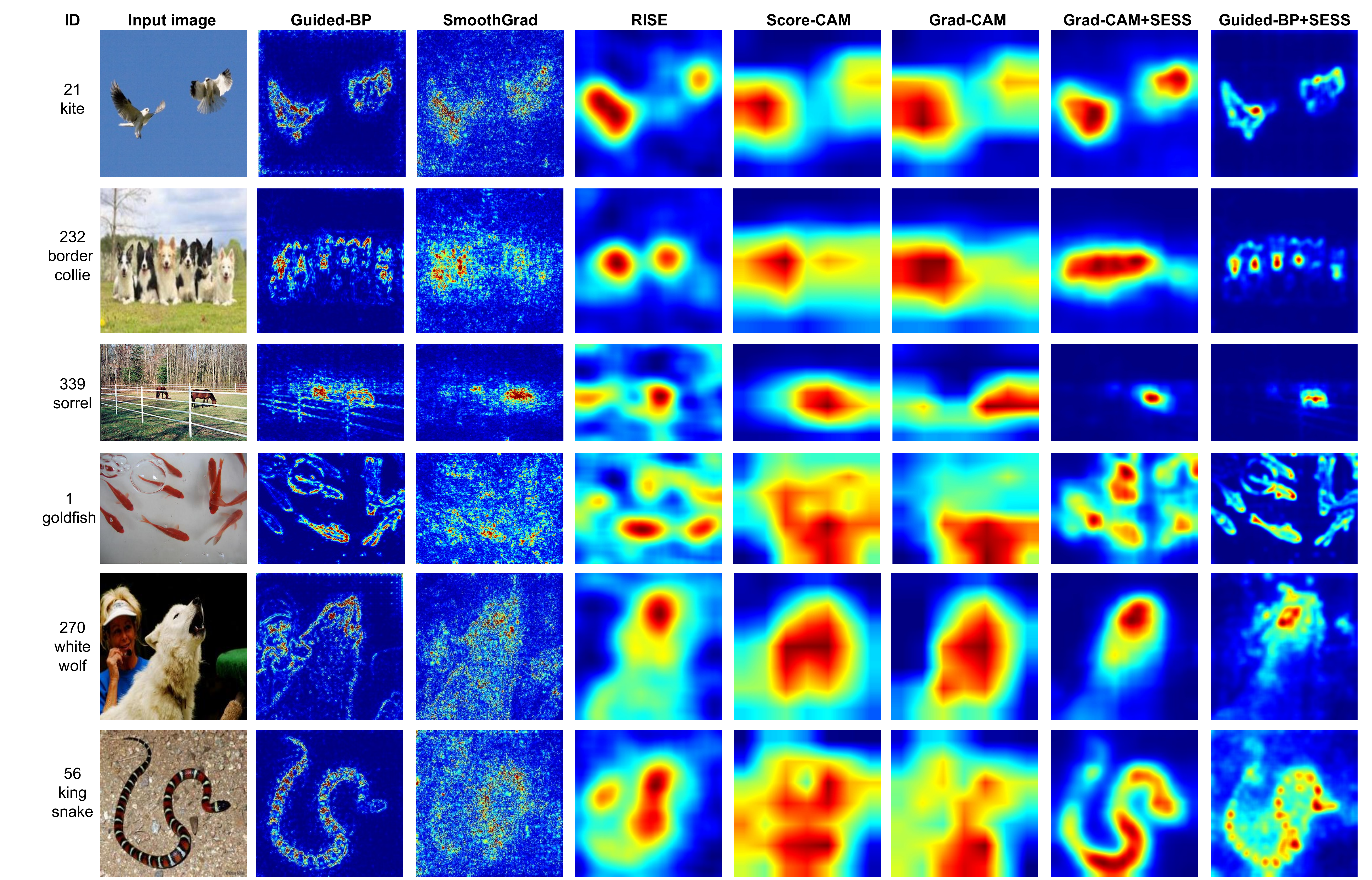}
	\caption{Qualitative comparison with SOTA saliency methods. Target class IDs and inputs images are shown in the first two columns. Later columns show saliency maps for (from left to right) Guided-BP, SmoothGrad, RISE, Score-CAM, Grad-CAM, Grad-CAM with SESS, and Guided-BP with SESS.}
	\label{fig:qual}
\end{figure*}

As shown in Fig. \ref{fig:qual}, visualisation results with SESS are more discriminative and contain less noise. SESS reduces noise and suppresses distractors from the saliency maps, while making the Grad-CAM maps more discriminative and robust to small scales and multiple occurrences.

\subsection{Running Time}
We calculated the average running time of Grad-CAM, Guided-BP and Group-CAM with/without SESS on a randomly selected set of 5000 images from the ImageNet-1k validation split. Since SESS's running time is decided by the pre-filter ratio and the number of scales, we calculated SESS's running time with two scales (6 and 12) and three pre-filtering ratios (0, $50\%$, $99\%$). For comparison, we also calculated the average running time of RISE \cite{petsiuk2018rise}, Score-CAM \cite{wang2020score} and XRAI \cite{kapishnikov2019xrai}. These experiments are conducted with an NVIDIA T4 Tensor Core GPU and four Intel Xeon 6140 CPUs. Results are given in Table \ref{tab:time}. With SESS, the average computation time increased, though it is substantially reduced by using a higher pre-filter ratio and a lower number of scales. Compared to perturbation-based methods, activation/gradient-based methods with SESS are efficient. For instance, in the worst case, SESS requires 16.66 seconds which is still over twice as fast as RISE and XRAI, which require more than 38 seconds for a single enquiry.

\begin{table}[!ht]
	\centering
	\begin{tabular}{c|c|c|c|c}
		\hline
		\multirow{3}{*}{\bf Method} & \multirow{2}{*}{\bf Without}     & \multicolumn{3}{c}{\bf With SESS}                                       \\ \cline{3-5} 
		& \bf SESS
		&
		\multicolumn{1}{l|}{\begin{tabular}[c]{@{}c@{}} \bf Pre-filter=0\%\\ \bf Scale=6/12\end{tabular}} &
		\multicolumn{1}{r|}{\begin{tabular}[c]{@{}c@{}} \bf Pre-filter=50\%\\ \bf Scale=6/12\end{tabular}} &
		\begin{tabular}[c]{@{}c@{}} \bf Pre-filter=99\%\\ \bf Scale=6/12\end{tabular} \\ \hline
		Grad-CAM  \cite{selvaraju2017grad}               & 0.03 & 1.23/3.96 & 0.84/2.36 & 0.42/0.70 \\ \hline
		Guided-BP  \cite{springenberg2014striving}        & 0.04 & 1.29/4.09 & 0.86/2.37 & 0.41/0.73  \\ \hline
		Group-CAM  \cite{zhang2021group}              &              0.13             & 4.42/16.66      & 2.54/8.54      &  0.51/0.92     \\ \hline
		RISE  \cite{petsiuk2018rise}             &              38.25             &  -      & -      &  -     \\ \hline
		Score-CAM  \cite{wang2020score}              &              2.47             &  -      & -      &  -     \\ \hline
		XRAI \cite{kapishnikov2019xrai}             &              42.17             &  -      & -      &  -     \\ \hline
	\end{tabular}
	\caption{Comparison of average run-time (in seconds) of a single query from a set of 5000 randomly selected images from the ImageNet-1K validation split.}
	\label{tab:time}
\end{table}

\subsection{Localisation Evaluation}
In this section, we evaluate SESS using the Pointing Game introduced in \cite{zhang2018top}. This allows us to evaluate the performance of the generated saliency maps on weakly-supervised object localisation tasks. The localisation accuracy of each class is calculated using $Acc = {\#Hits}/({\#Hits + \#Misses})$. $\#Hit$ is increased by one if the highest saliency point is within the ground-truth bounding box of the target class, otherwise $\#Misses$ is incremented. The overall performance is measured by computing the mean accuracy across different categories. Higher accuracy indicates better localisation performance.

We conduct the Pointing Game on the test split of PASCAL VOC07 and the validation split of MSCOCO2014. VGG16 and ResNet50 networks are used as backbones, and are initialised with the pre-trained weights provided by \cite{zhang2018top}. For implementation, we adopt the TorchRay \footnote{https://facebookresearch.github.io/TorchRay} library. Grad-CAM, Guided-BP and Group-CAM are chosen as base saliency methods. For efficiency and fair comparison, we set the pre-filter ratio to $99\%$ and the number of scales to 10. As per \cite{zhang2018top}, the results of both the ``all" and ``difficult'' sets are reported. The ``difficult" set includes images with small objects (covering less than 1/4 of the image) and distractors.  

Results are shown in Table \ref{tab:pg}. With SESS, all three methods achieved significant improvements, especially on the ``difficult" set. The average improvement for all cases is $11.2\%$, and the average improvement for difficult cases is $19.8\%$. Grad-CAM with SESS achieved SOTA results. The results further demonstrate that the multi-scaling and sliding window steps of SESS are beneficial when scale variance and distractors exist.

\begin{table}[!htb]
	\centering
	\begin{tabular}{c|c|c|c|c|c}
		\hline
		\multirow{2}{*}{\textbf{Method}} &
		\multirow{2}{*}{\textbf{SESS}} &
		\multicolumn{2}{c|}{\textbf{VOC07 Test (All/Diff)}} &
		\multicolumn{2}{c|}{\textbf{COCO Val. (All/Diff)}} \\ \cline{3-6} 
		&     & \multicolumn{1}{c|}{\textbf{VGG16}} & \textbf{ResNet50} & \multicolumn{1}{c|}{\textbf{VGG16}} & \textbf{ResNet50}     \\ \hline
		\multirow{2}{*}{Grad-CAM \cite{selvaraju2017grad}} &
		 &
		86.6/74.0 &
		90.4/82.3 &
		54.2/49.0 &
		57.3/52.3 \\ \cline{2-6} 
		& \checkmark & \bf 90.4/80.8      & \bf 93.0/86.1         & \bf 62.0/57.8               & \bf 67.0/63.2             \\ \hline
		\multirow{2}{*}{Guided-BP \cite{springenberg2014striving}} &
		 &
		75.9/53.0 &
		77.2/59.4 &
		39.1/31.4 &
		42.1/35.3 \\ \cline{2-6} 
		& \checkmark & 79.4/64.2      &    86.0/75.7               & 39.5/34.5               & 44.0/39.4 \\ \hline
		
				\multirow{2}{*}{Group-CAM \cite{zhang2021group}} &
		&
		80.2/64.9 &
		84.2/71.0 &
		47.4/41.1 &
		48.6/42.4 \\ \cline{2-6} 
		& \checkmark & 89.5/79.8      &    92.4/85.3               & 61.2/56.9               & 66.2/62.3 \\ \hline
		
		RISE \cite{petsiuk2018rise} &    & 86.9/75.1 & 86.4/78.8 & 50.8/45.3 & 54.7/50.0 \\\hline
		EBP \cite{zhang2018top} & & 77.1/56.6 & 84.5/70.8 & 39.8/32.8 & 49.6/43.9 \\\hline
		EP \cite{fong2019understanding} & & 88.0/76.1 & 88.9/78.7 &  51.5/45.9 & 56.5/51.5 \\\hline
	\end{tabular}
	\caption{Comparative evaluation on the Pointing Game \cite{zhang2018top}.}
	\label{tab:pg}
\end{table}

\section{Conclusion}
\label{sec:con}

In this work, we proposed SESS, a novel model and method agnostic extension for saliency visualisation methods. As qualitative results show, with SESS, the generated saliency maps are more visually pleasing and discriminative. Improved quantitative experimental results on object recognition and detection tasks demonstrate that SESS is beneficial for weakly supervised object detection and recognition tasks. 
\bibliographystyle{splncs04}
\bibliography{egbib}

\section{Supplementary Materials}

\subsection{More qualitative Results}
This section provided qualitative results related to the step size and weighted average.

\noindent{\bfseries{Weighed average}} In the fusion step, a weighted average is applied to ignore zero saliency values introduced by the calibration step. As Fig. \ref{fig:wa} shows, without the weighted average, some parts of the target object will be under activated. For example, near the tale of the snake and cat. The saliency values of those under activated regions are increased with a weighted average.

\begin{figure*}[!tbh]
	\centering
	\includegraphics[width=0.8\linewidth]{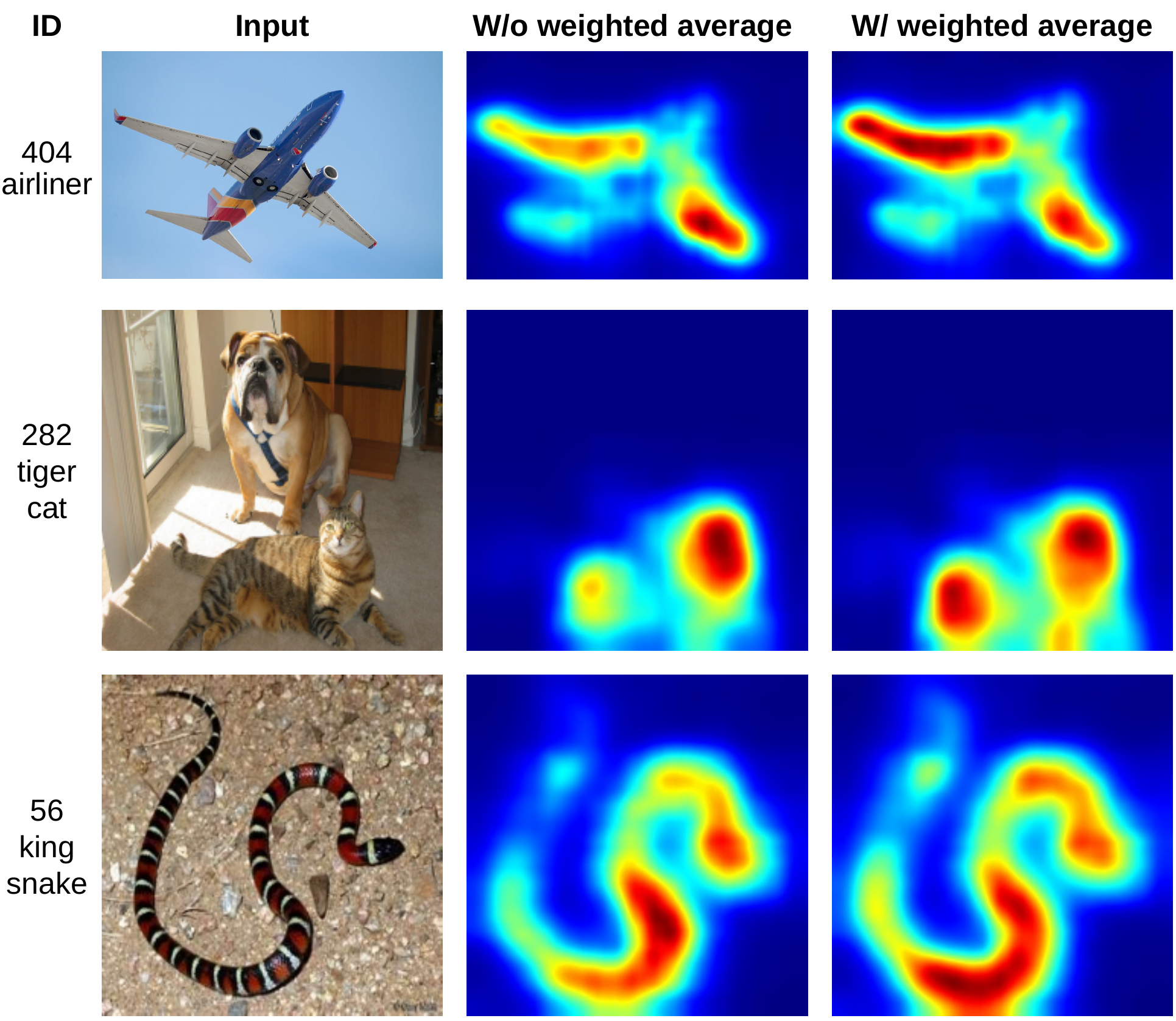}
	\caption{Impact of weighted average: The weighted average increases the saliency values of under activated regions.}
	\label{fig:wa}
\end{figure*}

\begin{figure*}[!tbh]
	\centering
	\includegraphics[width=0.8\linewidth]{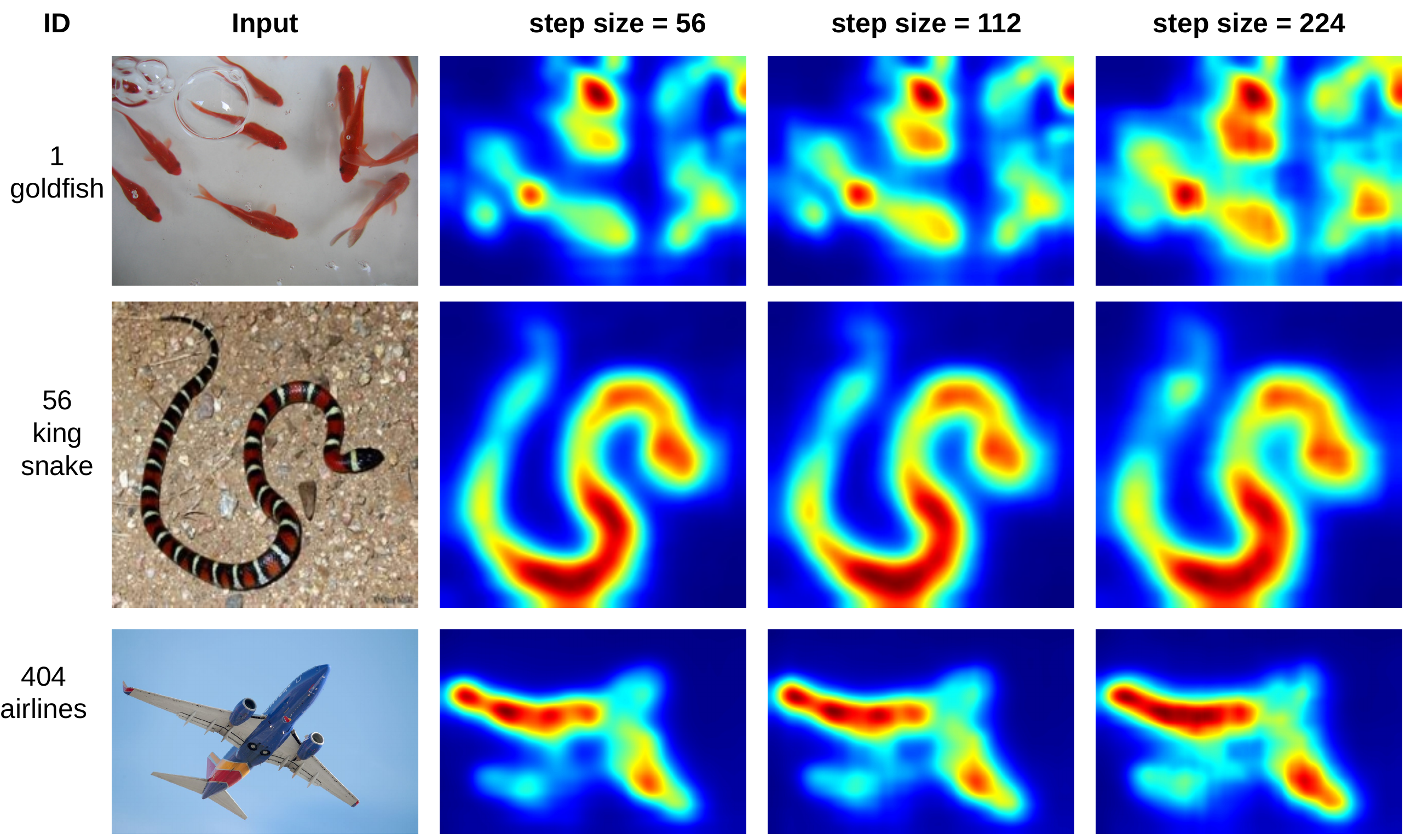}
	\caption{Impact of step-size: A larger step size reduces the over activated regions near the boundary of the object.}
	\label{fig:ss}
\end{figure*}

\noindent{\bfseries{Step-size}} In the default implementation of SESS, the step-size is set to 224 for efficiency. However, a smaller step size is beneficial for the generation of an accurate saliency map. As shown in Fig. \ref{fig:ss}, with a smaller step-size, the boundary of the target object is more accurate.

\subsection{Applications}
\begin{figure*}[!tbh]
	\centering
	\subfloat[][ResNet50 + Grad-CAM]{\includegraphics[width=0.6\linewidth]{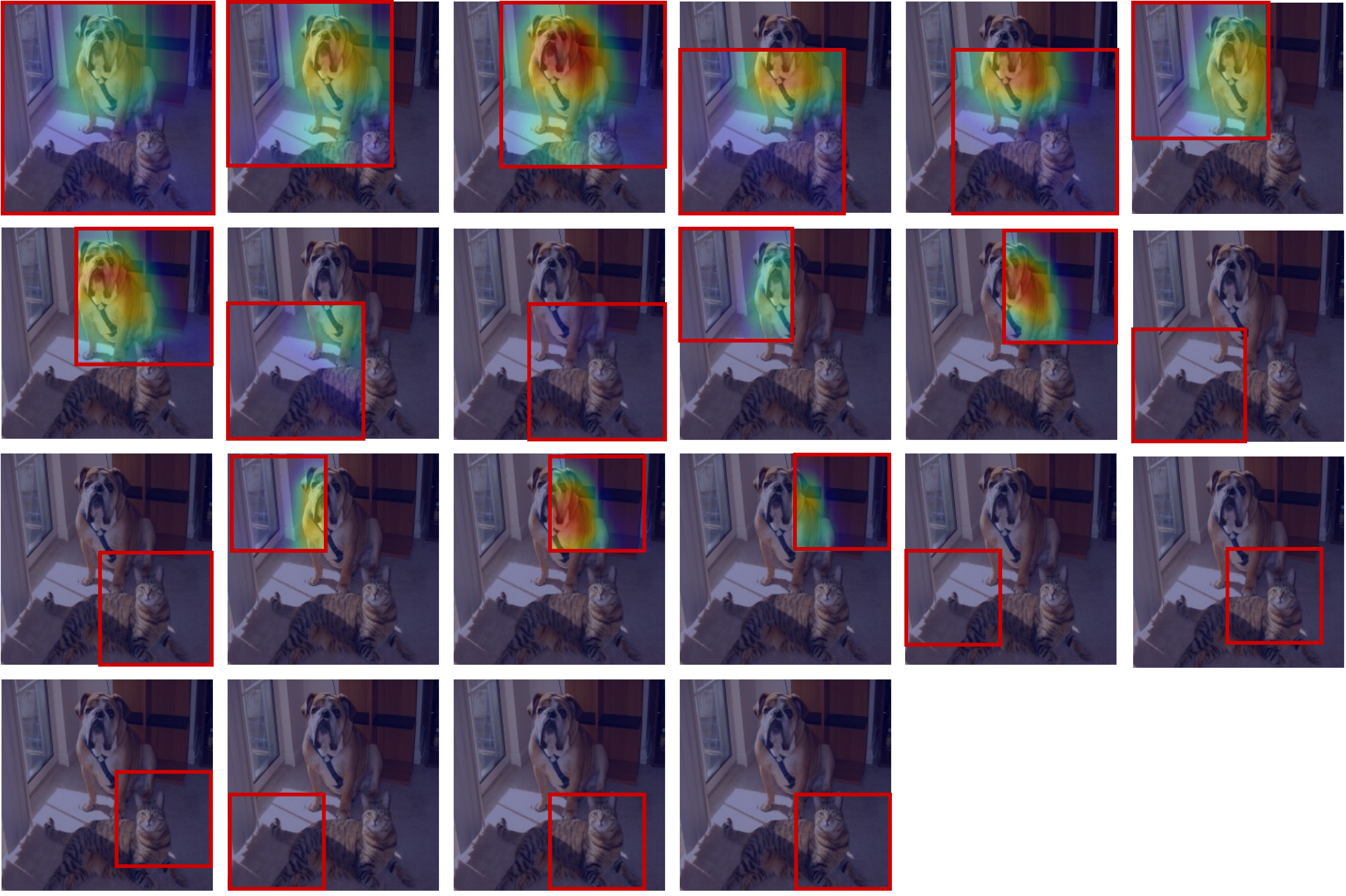}}
	
	\subfloat[][VGG-16 + Grad-CAM]{\includegraphics[width=0.6\linewidth]{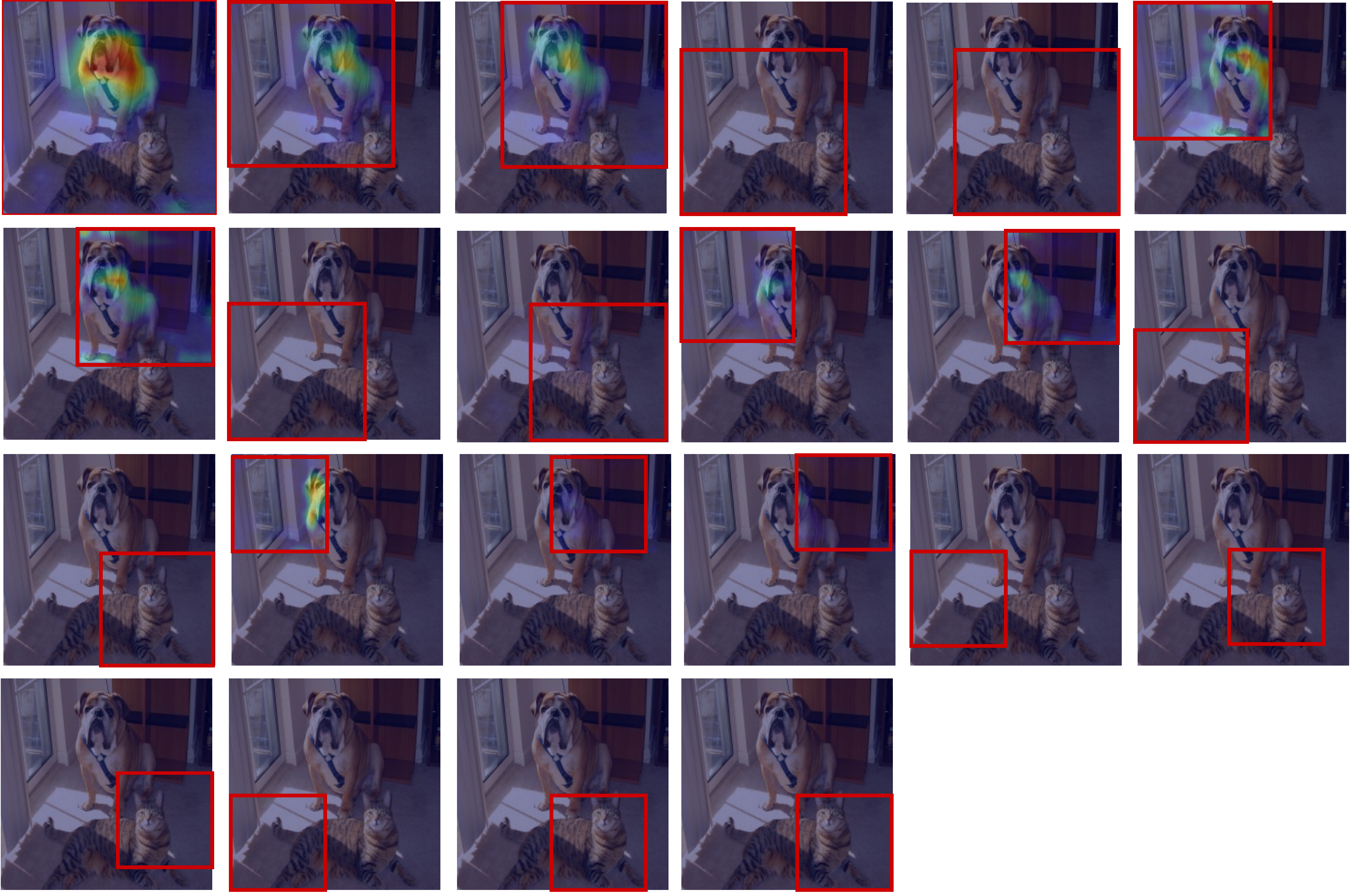}}
	
	\subfloat[][VGG-16 + Score-CAM ]{\includegraphics[width=0.6\linewidth]{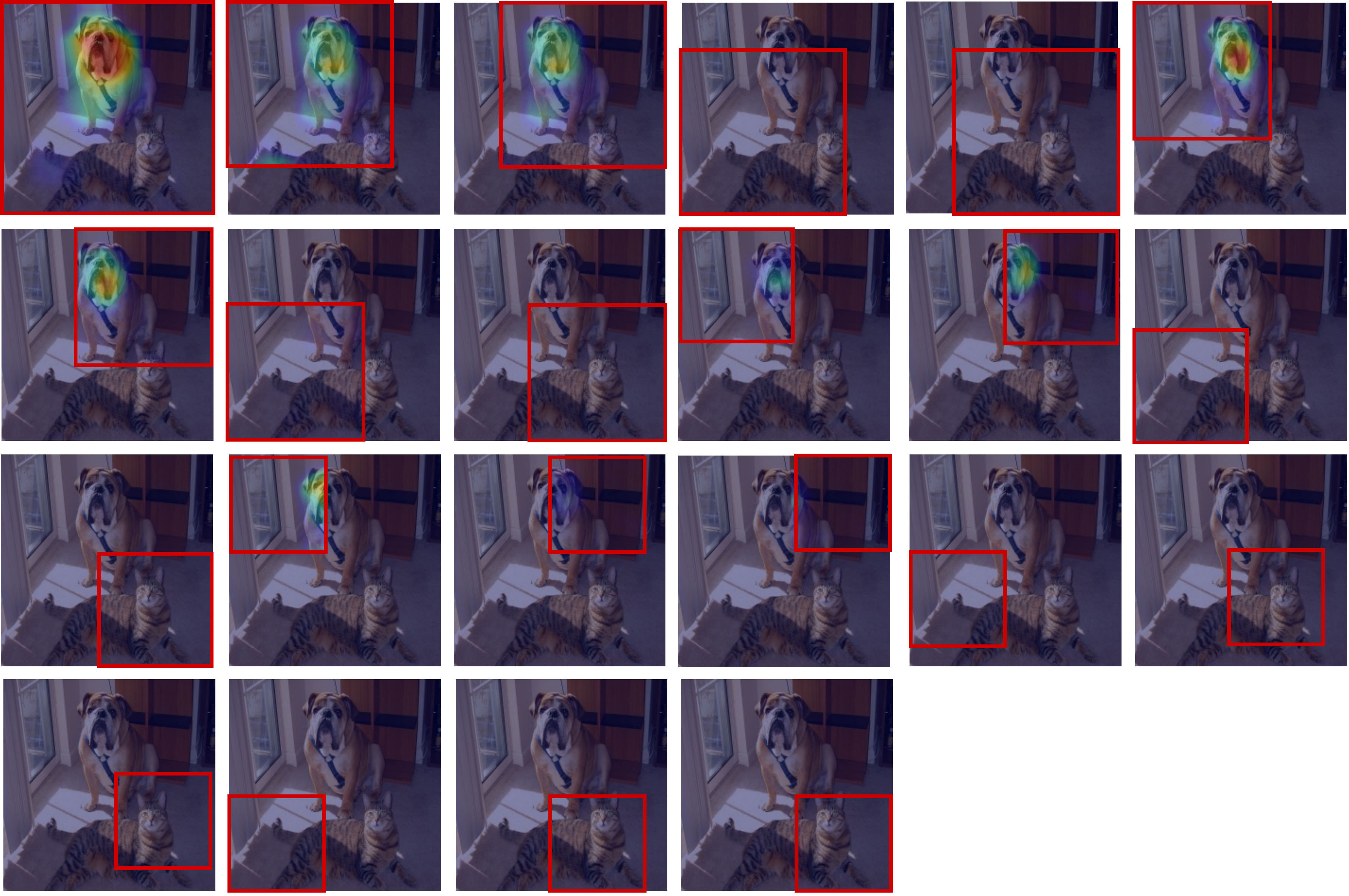}}
	\caption{Analysing DNN models and saliency visualisation methods with SESS. In this example the number of scales of SESS is set to 5. The red bounding box denotes the region from which the saliency is extracted. The target class id is 243 (Bull Mastiff).}
	\label{fig:ana}
\end{figure*}

SESS is also useful for analysing the DNN models and saliency visualisation methods. This can be done through visualising all extracted saliency maps in $L'$ as shown in Fig. \ref{fig:ana}. This visualisation shows: ResNet50 \cite{he2016deep} is more robust to scale and occlusion when compared to VGG-16 \cite{simonyan2014very}, and ScoreCAM is more robust to scale variance when compared to Grad-CAM.

\end{document}